\documentclass[conference]{IEEEtran}

\usepackage{times,enumerate} 
\usepackage[usenames,dvipsnames,svgnames,table]{xcolor}

\usepackage{subfloat}
\usepackage{hyperref}

\usepackage{graphicx}
\usepackage{setspace}
\usepackage{bbm}
\usepackage{mathdots,mathrsfs}
\usepackage{amssymb,latexsym,amsfonts,amsmath,cite,comment,relsize}
\usepackage{stmaryrd}
\usepackage{caption}
\usepackage[dvips]{psfrag}
\usepackage{setspace}
\usepackage{amsmath}
\usepackage{url}
\usepackage{soul}
\usepackage{blindtext}
\usepackage{graphicx}
\usepackage{lipsum}
\newcommand{\Sp}{\hspace{0.05cm}}

\usepackage{adjustbox}
\usepackage{tikz}
\usepackage{pgfplots}
\pgfplotsset{every tick label/.append style={font=\small}}

\usepackage{resizegather}

\usepackage{relsize}
\usetikzlibrary{patterns}

\usepackage{epsfig} 
\usepackage{subfig}
\usepackage{algorithm,algpseudocode}

\newtheorem{theorem}{Theorem}
\newtheorem{lemma}{Lemma}

\newtheorem{definition}{Definition}
\newtheorem{remark}{Remark}

\tikzstyle{vertex}=[circle, draw, inner sep=0pt, minimum size=6pt]

\usepackage[sans]{dsfont}
\usepackage[font={small,it}]{caption}

\newcommand{\suchthat}{\;\ifnum\currentgrouptype=16 \middle\fi|\;}

\newcommand{\R}{\mathbb{R}}
\newcommand{\Z}{\mathbb{Z}}

\newcommand{\Omg}{\mathbf{H}}
\newcommand{\mP}{\mathbf{\Sigma}}

\newcommand{\PD}{\mathcal{S}_{++}}

\title{\LARGE \bf
	Estimation with Fast Landmark Selection in  Robot Visual Navigation
}

\begin{document}
	
		\author{\authorblockN{Hossein K. Mousavi}
		\authorblockA{Department of Mechanical\\Engineering and Mechanics\\
			Lehigh University\\
			Bethlehem, Pennsylvania 18015\\
			Email: mousavi@lehigh.edu}
		\and
		\authorblockN{Nader Motee}
		\authorblockA{Department of Mechanical\\Engineering and Mechanics\\
			Lehigh University\\
			Bethlehem, Pennsylvania 18015\\
			Email: motee@lehigh.edu}}
	
	\maketitle
	\thispagestyle{empty}
	\pagestyle{empty}

	\begin{abstract} We consider the visual feature selection to improve the estimation quality required for the accurate navigation of a robot. We build upon a key property that asserts: contributions of trackable features (landmarks) appear linearly in the information matrix of the corresponding estimation problem. We utilize standard models for motion and vision system using a camera to formulate the feature selection problem over moving finite time horizons. A scalable randomized sampling algorithm is proposed to select more informative features (and ignore the rest) to achieve a superior position estimation quality. We provide probabilistic performance guarantees for our method. The time-complexity of our feature selection algorithm is linear in the number of candidate features, which is practically plausible and outperforms existing greedy methods that scale quadratically with the number of candidates features.  Our numerical simulations confirm that not only the execution time of our proposed method is comparably less than that of the greedy method, but also the resulting estimation quality is very close to the greedy method. 
	\end{abstract}


	\section{Introduction}
	
	Safe and robust navigation is one of the fundamental problems in robotics. The recent technological advances in the computing devices have opened up new opportunities and made several breakthroughs possible in this research area \cite{yang2018grand}. However, the research efforts to devise scalable algorithms for planning and perception during robot navigation is still immature. Applications of mobile robots will, at least for a few years, suffer from onboard computational and power limitations. Even if these physical or computational boundaries are pushed further away, the demand for agility and higher levels of autonomy always mandates us to execute onboard procedures in shorter periods of time. 
	
	One of the essential subproblems during robot navigation is to solve the localization, mapping, and visual odometry at an acceptable level of accuracy while spending a minimal amount of computational resources \cite{thrun2005probabilistic}. To achieve this goal, many researchers have investigated visual feature selection problem \cite{sim1999learning,sala2006landmark,lerner2007landmark,chen2008active,strasdat2009landmark,gorbenko2012problem,beinhofer2013robust,mu2017two}. The underlying idea is that depending on the current state of the robot and planned motion in the near future (i.e., the task), tracking certain features across a time horizon can be more informative than tracking other features. In other words, certain visual features may deserve more attention compared to the rest. In this regard, \cite{davison2005active} uses a greedy method to select a subset of pre-identified visual landmarks which facilitate the pose estimation of the robot.   
	In \cite{strasdat2009landmark}, the authors combine solving the simultaneous localization and mapping (SLAM) using the unscented Kalman filtering with reinforcement learning. Their approach generates policies that govern the feature selection. In \cite{mu2017two}, a two-stage methodology for measurement planning is discussed. The first stage is the selection of the subset of landmarks for observation, which is followed by the design of observation times for each feature. In \cite{lerner2007landmark}, the authors consider the problem of task-aware design of a subset of features such that an uncertainty metric is minimized. In \cite{carlone2018attention}, the authors consider a visual-inertial navigation problem. They analyze the problem of feature-selection for superior performance, where the design variable is the features that will be tracked during a fixed time-horizon. They use convex relaxations as well as the greedy method of selection and bring performance guarantees for the quality of estimation using the selected features. 
	
	In this paper, we aim at solving the problem of feature selection out of a given set of candidates in order to achieve superior estimation and anticipation during the navigation. The navigation setup consists of a robot that moves based on generated position estimates. The robot is assumed to use an onboard camera to (passively) track selected features in a fixed time horizon to improve the quality of the estimation. The underlying vision system model and the resulting feature selection problem are similar to those in  \cite{carlone2018attention}, while instead of using the greedy method and convex relaxations, we propose a random sampling algorithm for feature selection. In our approach, each available feature is assigned a sampling probability, which is used for finding a set of features via sampling. We demonstrate that the proposed method provides us with theoretical guarantees on the quality of the estimation. It turns out that time-complexity of our randomized sampling algorithm scales linearly with the number of available features, while time complexity of the greedy method of \cite{carlone2018attention} scales quadratically for the exact same problem. Numerical simulations confirm that the estimation quality using features provided by our randomized sampling is very close to the quality of estimation provided by the greedy method, while the required time to run our randomized sampling algorithm is significantly less than the greedy method. 
	
	The rest of this paper is organized as follows. In the remaining of this section, we introduce some notations that are used throughout the paper. In Section \ref{sec:problem}, we portray the big picture of the research problem that is being addressed in this paper. In Section \ref{sec:motionvision}, we discuss the details of the motion and vision models. Then, we define three estimation measures that quantify the quality of estimation based on the feature selection problem.  In Section \ref{sec:sampling}, we propose an algorithm for feature selection and conduct performance and time-complexity analysis for our approach. In Section \ref{sec:example}, a numerical case is explained, wherein we compare the performance of the proposed feature selection method against the totally random choice of features and the greedy method. Finally,  the final remarks are discussed in Section \ref{sec:conc}. {The proofs of the theoretical results   are given in the appendix}. 
	
	\noindent {\it Notations and Preliminaries:} The set of nonnegative integer and real numbers are denoted by $\Z_+$ and $\R_+$, respectively.  The vectors and matrices are denoted by lower-case and upper-case letters, respectively (e.g. $  x$ and $  X$). The identity matrix of size $n$ is denoted by $  I_n$. 
	The set of {positive definite} matrices of size $n$ is denoted by {  $\PD^n$}. The partial ordering on the cone of positive-semidefinite matrices is denoted by $\succ,~\succcurlyeq,~\prec$, and $\preccurlyeq$ operators. The block-diagonal matrix with diagonal elements $  X_1,\dots,  X_N$ is denoted by $\mathrm{diag}(  X_1,\dots,  X_N)$. For a set $S$, $|S|$ denotes its cardinality.   For a map $\mathbf g$, $\nabla_{  x} \mathbf g$ denotes the corresponding partial derivative.  A Gaussian random variable with mean vector $ {\mu}$ and covariance matrix $\Sigma$ is denoted by $\mathcal{N}( {\mu} ,  \Sigma)$.  $  X\otimes   Y$ denotes the Kronecker product of matrices $  X$ and $  Y$. The special orthogonal group   in $3$ dimensions is denoted by $\mathrm{SO}(3)$.

	\section{Problem Statement}\label{sec:problem}
	We formulate the feature selection problem for visual navigation of robots.  The spatial location of a robot at time $t \in \Z_+$ is denoted by $x_t \in \R^{3}$. For a given positive integer {$T$}, the vector of future states over the discrete time horizon $[ t,t+T ]=t,t+1,\dots,t+T$ is represented by   
	\begin{align*}
	\mathbf{x}_{t,T}:=\Big [x_t^T,~x_{t+1}^T,~\dots~,~x_{t+T}^T \Big ]^T \in \R^{3 (T+1)}. 
	\end{align*}
	Since robot motion creates uncertainty, having access to the statistics of $\mathbf{x}_{t,T}$ will help us measure quality of our prediction of robot whereabouts over the time horizon 
	\cite{carlone2018attention}. As it is shown in Subsection \ref{subsec:mov}, one can obtain 
	mean vector $\bar {\boldsymbol{\mu}}_{t,T} \in \R^{3(T+1)}$ and  covariance matrix $\bar {\mathbf {\Sigma}}_{t,T} \in {\PD^{3(T+1)}}$
	of  $\mathbf{x}_{t,T}$ under popular Gaussianity assumption. These quantities can be equivalently transformed into more relevant forms for the feature selection problem, namely, information vector and matrix, which are given by \cite{thrun2005probabilistic} 
	\begin{eqnarray} 
	\bar {\mathbf {b}}_{t,T}&=&\bar{\boldsymbol{\mu}}_{t,T}^T \hspace{0.08cm}\bar{\mathbf \Sigma}_{t,T}^{-1} \label{eq:barhdef}\\
	\bar {\mathbf H}_{t,T}& =&\,\bar {\mathbf \Sigma}_{t,T}^{-1} \label{eq:barbdef}.
	\end{eqnarray}
	
	There is a one-to-one correspondence between mean vector and covariance matrix and their counterparts information vector and information matrix. A striking property of the latter representation is that the contribution of each feature  (or landmark) to the information vector and matrix is fused linearly \cite{thrun2004simultaneous}. 
	Having the prior estimation parameters \eqref{eq:barhdef}-\eqref{eq:barbdef}, as it is shown in Subsection \ref{subsec:cameramodel}, the quality of estimation for $\mathbf x_{t,T}$ can be improved by fusing information of newly observed visual features using an onboard camera. The updated information matrix and vector are
	\begin{eqnarray} 
	\mathbf H_{t,T}(\Theta_t) &=&\bar {\mathbf H}_{t,T} +\sum_{f \in \Theta_t} \mathbf {H}_{t,T}^f \label{eq:additiveone}\\
	{\mathbf b}_{t,T}(\Theta_t)&=&\bar {\mathbf b}_{t,T} +\sum_{f \in \Theta_t} \mathbf b_{t,T}^f \label{eq:additivetwo}
	\end{eqnarray}
	in which $\mathbf b_{t,T}^f$ and $\mathbf {H}_{t,T}^f$ are contributions of feature $f$ to the overall information matrices of the estimation problem. The set of all identifiable features (landmarks) at time $t$, which can be triangulated using multiple frames over the time horizon $[t,t+T ]$, is denoted by $\Theta_t$. Suppose that  $|\Theta_t|=N_t$ is assumed to be large. 
	
	Tracking a large number of features (landmarks) for accurate navigation usually requires substantial onboard computational power \cite{sala2006landmark}. As a result, a desirable navigation objective is to select and track a small subset of features that are more informative, while providing an acceptable estimation quality. Suppose that robot is only capable of tracking at most $q$, which is comparably less than $N_t$, features during the horizon. 
	
	\vspace{0.1cm}
	\begin{definition}
		A map  $\rho:~\PD^n \rightarrow \R$ is called monotone decreasing if  $ {X} \preceq  {Y}$  implies  $\rho( {X})\geq \rho( {Y})$.
	\end{definition}
	\vspace{0.1cm}
	
	Then, the feature selection problem can be formulated as 
	\begin{flalign}
	&\underset{\Phi_t \subset \Theta_t }{\mathrm{minimize}}~~\rho \big( \mathbf H_{t,T}(\Phi_t) \big) & & \label{optim-1}\\
	&{\mathrm{subject~to:}} ~~~|\Phi_t| \leq q & &  \label{optim-2}
	\end{flalign}
	where $\rho:~\PD^{3(T+1)} \rightarrow \R$ is a monotone decreasing map that measures the estimation quality.    
	
	The optimization problem \eqref{optim-1}-\eqref{optim-2} is combinatorial and usually NP-hard.  The \emph{research problem} is to propose a scalable algorithm that provides   solutions for \eqref{optim-1}-\eqref{optim-2} with  performance guarantees.

	\section{Models for Robot Motion and Vision System} \label{sec:motionvision}
	
	In the following, we discuss details of models for the motion of the robot and the onboard vision system.
	
	\subsection{Statistics of Robot Motion} \label{subsec:mov}
	
	The goal is to calculate information vector and matrix of $\mathbf{x}_{t,T}$ when dynamics of robot evolves over time horizon $[ t,t+T ]$. To achieve this, we utilize a model that is inspired by the dynamic model analyzed in \cite{thrun2004simultaneous}. Suppose that dynamics of the robot's position  is governed by     
	\begin{align}\label{eq:robotdynamics}
	x_{\tau}=\mathbf{g}(x_{\tau-1},  u_\tau)+ \delta_\tau
	\end{align}
	for all $\tau \in [ t+1,t+T ]$, where  {$\mathbf g: \R^3 \times \R^3 \rightarrow \R^3$} is a (possibly nonlinear) known map, where $  u_{\tau}$ in the control  command at time $\tau$, and $ \delta_{\tau} \sim \mathcal{N} (0,   \Lambda_{\tau})$ is a temporally independent random process that captures the  aggregate effect of all uncertainties induced by the robot motion. It is assumed that a feedback control law with the following structure is given  
	\begin{align}\label{eq:controllaw}
	u_{\tau}=\mathbf h(  u_{\tau}^{\mathrm{ref}},  x_{\tau-1})
	\end{align}
	that ensures the robot with dynamics \eqref{eq:robotdynamics} tracks a reference  path, at least in the absence of uncertainties, with some desired accuracy. The command $u_{\tau}^{\mathrm{ref}}$ may have to be pre-filtered to enhance the tracking performance.
	Since robot's position $x_{\tau-1}$  is a random variable and its true value is unknown, we use its mean value $\bar \mu_{\tau-1}$ as a faithful estimate of its value in \eqref{eq:controllaw} to obtain     
	\begin{align}\label{eq:controllawhat}
	{\bar{u}_{\tau}}=\mathbf h\left ( u_{\tau}^{\mathrm{ref}},\bar \mu_{\tau-1} \right ). 
	\end{align}
	In presence of uncertainties, the  trajectory of the closed-loop system \eqref{eq:robotdynamics}-\eqref{eq:controllawhat}  will fluctuate around the reference path and the  
	tracking quality will depend on  the quality of estimation $\bar \mu_{\tau-1}$. The control mechanism \eqref{eq:controllawhat} is merely using the initial statistics of position of the robot. In the next subsection, we show that incorporating new information obtained from observing features (landmarks) will help us improve the estimation quality, which in turn will improve the path tracking quality.   
	

	Suppose that the current, i.e., before accounting for the dynamics of the robot, pose estimates for ${x}_{t}$ is described  by mean vector ${\mu}_{t}$ and covariance matrix $\Sigma_{t}$. For the time step starting at $\tau=t$ in the horizon, let us set $\bar {\mu}_{t}={\mu}_{t}$ and $\bar {  \Sigma}_{t}=  {  \Sigma}_{t}$. For the next steps, we define the composed map
	\begin{align}
	\mathbf f(  x, \mu,  u):=\mathbf g \big(  x,\mathbf h(  u,\mu) \big). 
	\end{align}
	Then, upon linearizing the dynamics of the system at working point $(x,\mu,u)= (\bar{\mu}_{\tau-1}, \bar{\mu}_{\tau},u_{\tau}^{\mathrm{ref}})$  with respect to $x$, we get
	\begin{align} \label{eq:xdynamics}
	x_{\tau} \hspace{0.08cm} \approx \hspace{0.08cm} \bar \Delta_{\tau} +A_{\tau} \left (  {x}_{\tau-1}-  \bar{{\mu}}_{\tau-1} \right )+\delta_\tau
	\end{align}
	in which vector $\bar \Delta_{\tau}$ and $A_{\tau}$  are given by 
	\begin{eqnarray}
	\bar{ {\Delta}}_{\tau} & := & \mathbf g \big(  \bar{{\mu}}_{\tau-1},\mathbf h( u_{\tau}^{\mathrm{ref}},  \bar {  \mu}_{\tau-1}) \big)\\
	A_{\tau} & := & \nabla_{  x}\, \mathbf f(  \bar{{\mu}}_{\tau-1},  \bar{{\mu}}_{\tau-1}, u_{\tau}^{\mathrm{ref}})
	\end{eqnarray}
	for all $\tau \in [t+1,t+T ]$.
	
	\begin{lemma} \label{lem:one} By setting $\bar {\mu}_{t}={\mu}_{t}$ and $\bar { \Sigma}_{t}=  {  \Sigma}_{t}$, the mean and covariance of $\mathbf x_{t,T}$ is given by  
		\begin{eqnarray}
		\bar {\mathbf \Sigma}_{t,T} & = &
		\begin{bmatrix} 
		{  \bar {  \Sigma}_t} & \bar { \Sigma}_{t,t+1} &\dots  & \bar { \Sigma}_{t,t+T} \\
		\bar { \Sigma}_{t,t+1}^T  & \bar { \Sigma}_{t+1} & \dots  &\bar { \Sigma}_{t+1,t+T}  \\
		\vdots & \vdots & \ddots & \vdots \\
		\bar { \Sigma}_{t,t+T}^T & \bar { \Sigma}_{t+1,t+T}^T & \dots & \bar { \Sigma}_{t,T}
		\end{bmatrix} \label{eq:updated}  \\
		\bar{\boldsymbol{\mu}}_{t,T}  & = & \Big [ \bar \mu_{t}^T \,~~\, \bar \mu_{t+1}^T\,~~\dots~~\,\bar \mu_{t+T}^T\ \Big ]^T  \label{eq:updatedd}
		\end{eqnarray}
		where      
		\begin{eqnarray*}  
			\bar { \Sigma}_{\tau} &=&  { A}_\tau  \bar { \Sigma}_{\tau-1}    { A}_\tau^T +   \Lambda_\tau    \\
			\bar {\mathbf  \mu}_\tau & = &  \bar { \Delta}_\tau    
		\end{eqnarray*}    
		for every instant $\tau \in [t+1,t+T ]$ and 
		\[
		\bar { \Sigma}_{\tau_1,\tau_2}=\left ( \prod_{i=1}^{\tau_2-\tau_1} { A}_{\tau_2-i-1} - I \right)   \bar { \Sigma}_{\tau_1}   \]
		for all $\tau_1,\tau_2 \in [t,t+T ]$ with $\tau_1<\tau_2$.     
	\end{lemma}
	
	We can substitute \eqref{eq:updated} and \eqref{eq:updatedd} into \eqref {eq:barhdef} and \eqref{eq:barbdef} to calculate  information vector $\bar {\mathbf {b}}_{t,T}$ and  matrix $\bar {\mathbf H}_{t,T}$. In the next subsection, it is shown that these  vectors and matrices will be updated upon receipt of certain information about  the observed features over the time horizon $[t,t+T]$.

	\subsection{ Camera Model for Feature Tracking} \label{subsec:cameramodel}
	
	
	We employ the observation model proposed by  in \cite{carlone2018attention} for an onboard camera. For every $\tau \in [t,t+T]$, let us denote  orientation of the robot by rotation matrix $R_\tau \in \mathrm{SO}(3)$, orientation of the camera with respect to the robot by  rotation matrix $R_{\mathrm{c}} \in \mathrm{SO}(3)$,   translation of the camera with respect to the robot pose by $x_{\mathrm{c}} \in \R^3$,  the unit vector corresponding to pixel measurement of feature $f \in \Theta_t$ at time $\tau$ by ${u_{\tau,T}^f} \in \R^3$, and  the position vector of the feature by $y_f\in \R^3$.  We recall that  the corresponding skew-symmetric matrix induced by ${u_{\tau,T}^f}$ satisfies 
	\[ U_{\tau,T}^f \hspace{0.05cm}  v = {u_{\tau,T}^f} \times   v   \]
	for all $ v \in \R^3$. As it is discussed in \cite{carlone2018attention}, one may reasonably assume that the observation vector in the image and its counterpart in real world are  collinear (i.e., parallel). However, due to existence of noise in the process, one may consider a disrupted version of this assumption by considering the following noisy observation model 
	\begin{align} \label{eq:cameramodel1}
	U_{\tau,T}^f 
	\left ( \left (   R_{\tau}   R_{\mathrm{c}} \right )^T 
	\left (   y_f -\left (   x_\tau + R_{\tau}   x_{\mathrm{c}}  \right ) \right ) \right ) =  {\eta}_{\tau,T}^f,
	\end{align}
	where $\eta_{\tau,T}^f \sim \mathcal{N}(  0,\sigma^ 2  {I}_3)$. The observation model \eqref{eq:cameramodel1} can be rewritten as 
	\begin{align} 
	z_{\tau,T}^f &=  U_{\tau,T}^f  \left (  R_{\tau} R_{\mathrm{c}} \right )^T (  x_\tau -  y_f)+ {\eta}_{\tau,T}^f \label{z-model}    
	\end{align}
	with  $ z_{f,\tau}:=- U_{f,\tau}      R_c^T   x_c = U_{f,\tau} ^T     R_c^T   x_c$. The camera takes  one frame at every time instant over time horizon $[t,t+T]$. With knowledge of planned motion (i.e., location and orientation) for robot over the time horizon,  suppose that robot is capable of running forward simulations to determine a feature will be visible in $n_f$ frames out of all $T+1$ frames over the time horizon. By considering relation \eqref{z-model} for such  visible features, one can stack all these equations  and write them in more compact form
	\begin{align}
	\mathbf z_{t,T}^f=  \mathbf F_{t,T}^f \mathbf x_{t,T}+\mathbf E_{t,T}^f   y_f +  \boldsymbol{\eta}_{t,T}^f,
	\end{align}
	for some appropriate  matrices $\mathbf F_{t,T}^f$ and $\mathbf E_{t,T}^f$. A given apriori information matrix $\bar {\mathbf H}_{t,T}$, which  is obtained from \eqref{eq:updated}, can be updated by fusing information of a visible feature $\{f\}$ according to the following rule \cite{carlone2018attention}
	\begin{align}
	\mathbf H_{t,T}(\{f\})=\bar {\mathbf H}_{t,T} +\mathbf {H}_{t,T}^f,
	\end{align}
	where the linearly added information matrix is given by 
	\begin{align*}\label{eq:hdefine}
	&\mathbf {H}_{t,T}^f=   \\
	&\begin{adjustbox}{max width=240 pt}
	$ \sigma^{-2} \Big( \left(\mathbf F_{t,T}^f \right)^T   \mathbf F_{t,T}^f -  \left(\mathbf F_{t,T}^f \right)^T   \mathbf E_{t,T}^f \left( \left(\mathbf E_{t,T}^f \right)^T   \mathbf E_{t,T}^f \right)^{-1} \left(\mathbf E_{t,T}^f \right)^T   \mathbf F_{t,T}^f \Big)$ \end{adjustbox}.  
	\end{align*}
	This additive property of the information matrix can be verified by application of the Bayes law \cite{thrun2004simultaneous} together with the Schur complement  \cite{carlone2018attention}. A similar treatment allows us to derive the following  update rule for the information vector.
	\begin{lemma}\label{lem:updateb}
		The information vector of  $\mathbf x_{t,T}$  upon tracking  feature $f \in \Theta_t$ is updated according to 
		\begin{align}
		{\mathbf b}_{t,T} (\{f\})=\bar {\mathbf b}_{t,T}  + \left (\mathbf B_{t,T}^f \hspace{0.05cm} \mathbf z_{t,T}^f \right )^{T}, 
		\end{align}
		where matrix $\mathbf B_{t,T}^f$ is given by 
		\begin{align*} \label{eq:bdefine}
		&\mathbf B_{t,T}^f:=\\
		& \begin{adjustbox}{max width=240 pt}
		$
		\sigma^{-2} \left (\left(\mathbf F_{t,T}^f \right)^T    -  \left(\mathbf F_{t,T}^f\right)^T   \mathbf E_{t,T}^f \left ( \left(\mathbf E_{t,T}^f \right)^T   \mathbf E_{t,T}^f \right)^{-1} \left(\mathbf E_{t,T}^f \right)^T     \right).
		$
		\end{adjustbox} 
		\end{align*}
	\end{lemma}
	
	\begin{figure}
		\begin{center}
			\includegraphics[width=6.0cm]{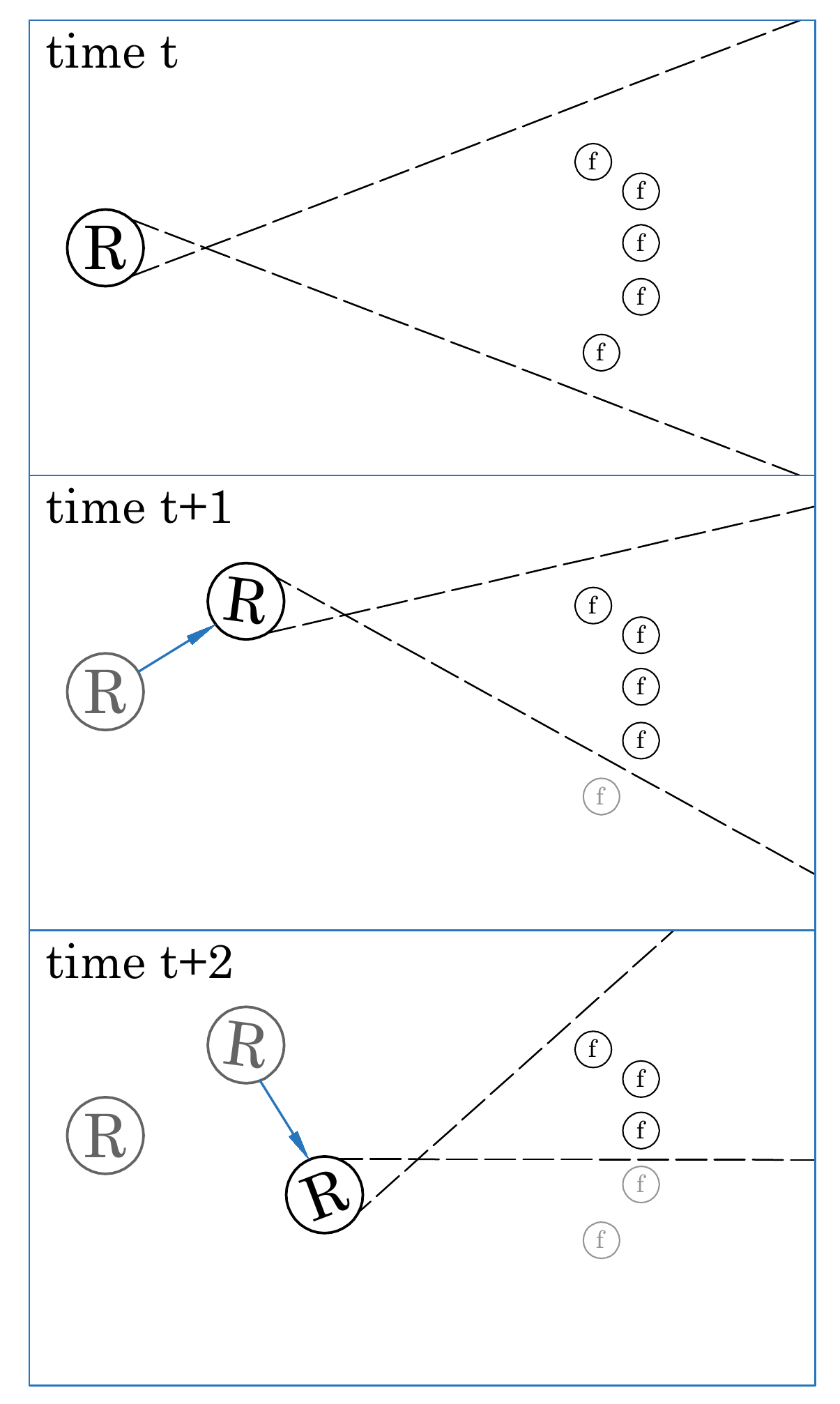}
			\caption{The schematic of the motion and vision model adopted in this paper at three consecutive snapshots. The location of the robot is denoted by letter $R$, which moves and changes its orientation across three frames. The features are denoted by letter $f$. As a result of this movement, the visible features (those in between the dashed lines) in each frame will vary. Set $\Theta_t$ will consist of all features that can be triangulated during this horizon. } 
			\label{fig:nav}
		\end{center}
	\end{figure}
	\vspace{1mm}
	Since contributions of different features are independent of each other,  for a selected subset of features   $\Phi_t \subset \Theta_t$, one can verify that the updates to the information matrix and vector upon the choice of these features are given by
	\begin{eqnarray} 
	\mathbf H_{t,T}(\Phi_t) & = & \bar {\mathbf H}_{t,T} +\sum_{f \in \Phi_t} \mathbf {H}_{t,T}^f, \label{eq:additiveone}\\
	{\mathbf b}_{t,T}(\Phi_t) & = & \bar {\mathbf b}_{t,T} +\sum_{f \in \Phi_t} \left (\mathbf B_{t,T}^f \mathbf z_{t,T}^f \right )^{T}.\label{eq:additivetwo}
	\end{eqnarray}
	The corresponding mean vector and covariance matrix for the estimation problem can be calculated through 
	\begin{eqnarray}  
	{\mathbf \Sigma}_{t,T}(\Phi_t) & =& {\mathbf H}_{t,T}  (\Phi_t) ^{-1}, \\
	\boldsymbol{\mu}_{t,T} (\Phi_t)^T & =& {\mathbf {b}}_{t,T}(\Phi_t) \hspace{0.05cm} {\mathbf H}_{t,T}  (\Phi_t)   ^{-1}.  
	\end{eqnarray}
	
	In Fig. \ref{fig:nav}, we illustrate the essence of the motion and vision model described in this section.  In order to specify the set of available features $\Theta_t$ for tracking, the robot needs to determine which features can be triangulated and which ones will result in invertible information matrices $(\mathbf E_{t,T}^f)^T   \mathbf E_f$ (cf.  \cite{carlone2018attention}).

	\subsection{Estimation Measures}
	
	Given a subset of trackable features $\Phi_t$, we can quantify the quality of resulting estimation in various meaningful ways.  
	
	\noindent  \textit{(i) Variance of the Error: } Given the covariance matrix, the variance of error equals the sum of the variances of all scalar components of vector $\mathbf x_{t,T}$. This measure can be characterized as 
	\begin{align}
	\rho_v(\mathbf H_{t,T}(\Phi_t)):=\mathrm{Tr}(\mathbf H(\Phi_t)^{-1})=\mathrm{Tr}(\mP(\Phi_t)).
	\end{align}
	
	\noindent \textit{(ii) Differential Entropy of the Estimation Error: } It is known that the differential entropy of a multivariate Gaussian random variable with  covariance $\mP$ is 
	\begin{align*}
	h = \frac{1}{2} \log (\det (\mP)) + \frac{n}{2} \big(1+ \log(2 \pi) \big).
	\end{align*}
	This measure quantifies the uncertainty volume of the estimation error, which is given by  
	\begin{gather*}
	\rho_e(\mathbf H_{t,T}(\Phi_t))=\log (\det (\mP(\Phi_t))) =-\log (\det (\Omg_{t,T}(\Phi_t))),  
	\end{gather*}
	
	\noindent  \textit{(iii) Spectral Variance: } Let us consider the eigen-space of the largest eigenvalue of the covariance matrix of the estimator. This is the subspace   across which the estimation is less accurate than the rest of the directions. Thus, we can use the following estimation measure
	\begin{gather*}
	\rho_\lambda(\mathbf H_{t,T}(\Phi_t))=\lambda_{\max} (  \mP(\Phi_t)) = \lambda_{\min} (  \Omg_{t,T}(\Phi_t))^{-1}.  
	\end{gather*}
	

	All these measures are monotonically decreasing. They are also spectral functions, i.e., they only depend on the eigenvalues of the information or covariance matrices. Therefore, having lower and upper bounds for the covariance matrix can be potentially useful to obtain similar bounds for these estimation measures. Measures {\it (ii)} and {\it (iii)} have been also discussed in \cite{carlone2018attention}.     
	
	\section{Feature Selection via Randomized Sampling}  \label{sec:sampling}
	
	We propose a scalable algorithm that provides feasible solutions for \eqref{optim-1}-\eqref{optim-2} with provable performance bounds.   
	
	\subsection{Leverage Scores and Induced Probabilities}
	
	Each available feature $f \in \Theta_t$ is assigned some nonnegative numbers, which are so-called leverage scores, that are closely related to the notion of effective resistances in graph sparsification problem  \cite{spielman2011graph}. The 
	maximal information matrix of $\mathbf x_{t,T}$,  over the cone of positive-definite matrices, corresponds to the case where all features are employed in the estimation process. This matrix is given by  
	\begin{align}\label{eq:maximinformation}
	{\mathbf {H}}_{t,T}(\Theta_t)=\bar {\mathbf H}_{t,T} 
	+\sum_{ f \in  {\Theta}_t } \mathbf{H}_{t,T}^f. 
	\end{align}
	Therefore, for any $\Phi_t \subset\Theta_t$, it holds that 
	\begin{align}
	\Omg_{t,T}(\Phi_t) \preceq  {\mathbf {H}}_{t,T}(\Theta_t). 
	\end{align}
	Hence, for every monotone decreasing map  $\rho:\mathcal{S}^{3(T+1)}_{++} \rightarrow \R$, it follows that 
	\[
	\rho(\mathbf H_{t,T}(\Theta_t))  \leq     \rho(\mathbf H_{t,T}(\Phi_t)). 
	\]
	Using the maximal matrix, we define 
	\begin{align}
	\bar {\mathbf H}_{t,T}^f:=\dfrac{1}{N_t}\bar \Omg_{t,T}+  \Omg_{t,T}^f. 
	\end{align}
	for every $f \in \Theta_t$ with $|\Theta_t|=N_t$.

	\begin{definition}
		For a given set of features $\Theta_t$, the   leverage scores are  nonnegative numbers that are defined by
		\begin{equation}\label{resistent.def}
		r_f := \mathrm{Tr}\left (  {\mathbf {H}}_{t,T}(\Theta_t)^{-1}\bar {\mathbf H}_{t,T}^f \right )  
		\end{equation}
		for every feature $f \in \Theta_t$.
	\end{definition}
	
	One can associate a probability  mass  function denoted by 
	$
	\pi: \Theta_t \rightarrow [0,1]
	$
	to  elements of $\Theta_t$ by setting 
	\begin{equation}
	\pi(f) =\pi_f  = \dfrac{r_f }{n},\label{prob-elements}
	\end{equation}
	where $n=3(T+1)$ is the dimension of $\bar \Omg_{t,T}$. 
	The resulting function is a well-defined probability mass function as we have 
	\[
	\sum_{f \in \Theta_t} \Sp \pi(f ) \Sp =\Sp \frac{1}{n} \Sp \sum_{f \in \Theta_t }  \Sp  {\rm Tr} \left  (  {\mathbf {H}}_{t,T}(\Theta_t)^{-1}  \bar {\mathbf H}_{t,T}^f \right )=1.
	\]

	\subsection{Sampling Algorithm}
	
	We describe details of our algorithm to select a subset of the most informative features over each moving time horizon. This algorithm is inspired by the graph sparsification using effective resistances \cite{spielman2011graph,batson2013spectral}, where the goal is to construct a sparse weighted graph based on a given weighted graph by ensuring that the Laplacians of the two graphs stay spectrally close to each other.      
	A variation of this method appears in  \cite{mousavi2018space} for the case of rank-one selected matrices when the constant term $\bar {\mathbf H}_{t,T}$ is zero. 
	
	The steps of our method are given in Algorithm \ref{alg:execution}. First, we iteratively and independently sample a feature from $\Theta_t$ with replacement for $q$ iterations. This sampling takes place according to probability mass function $\pi$, which is defined by \eqref{prob-elements}. The sampled feature $f$ is added to   $\Phi_t$ provided that it has not been sampled before. At the end of the procedure, set of selected features $\Phi_t$ will have at most $q$ elements.   

	\begin{remark}
		This randomized feature selection algorithm can be applied independently of the model of robot motion which was described in Section \ref{subsec:mov}. For instance, our feature-selection approach can replace the feature selection routines in the inertial-visual navigation setup described in \cite{carlone2018attention} that are based on the greedy method and convex relaxations. 
	\end{remark}

	\begin{algorithm}[t]
		\caption{Randomized Feature Selection }
		\label{alg:execution}
		\begin{algorithmic}  
			\algnewcommand\algorithmicinitz{\textbf{initialize:}}
			\algnewcommand\Init{\item[\algorithmicinitz]}
			\renewcommand{\algorithmicrequire}{\textbf{input:}}
			\renewcommand{\algorithmicensure}{\textbf{output:}}
			\Require initial information matrix $\bar {\mathbf H}_{t,T}$\\ ~~~~~~set of available features $ \Theta_t$, number of samples $q$
			\Ensure  selected features $\Phi_t$,   information matrix $\mathbf H_{t,T}$  
			\Init $ \Phi_t=\varnothing$,  $\mathbf H_{t,T}=\bar{\mathbf H}_{t,T}$
			
			\For {$k=1$ to $q$}
			\State sample a feature from $\Theta_t$ using distribution $\pi$~$\rightarrow f$
			\State select the corresponding matrix 
			$$
			\mathbf H \leftarrow   {\mathbf H} _{t,T}^f 
			$$
			

			\If {$f \notin \Phi_t$,} 
			\State add $f$ to $\Phi_t$  
			\State update the information matrix:
			$$ \mathbf H_{t,T}  \leftarrow \mathbf H_{t,T} +\mathbf H $$
			\EndIf
			\vspace{0.05cm}
			\EndFor
			
		\end{algorithmic}
	\end{algorithm} 
	
	%
	%
	%
	%
	
	\subsection{Performance Guarantee}
	
	It turns out that Algorithm \ref{alg:execution} provides us with an information matrix that is a constant-factor approximation to the maximal information matrix ${\mathbf {H}}_{t,T}(\Theta_t)$ given by \eqref{eq:maximinformation}.   
	

	\begin{theorem} \label{theorem:sparse}
		For a given parameter $\epsilon \in (0,1)$, suppose that Algorithm \ref{alg:execution} is executed with a  fixed $q=O(n \log n/\epsilon^2)<N_t$. Then, the resulting information matrix, see \eqref{eq:additiveone}, based on the resulting set of features $\Phi_t$,  satisfies 
		\begin{align}\label{eq:spectralbound}
		\Omg_{t,T}(\Phi_t) \succeq \dfrac{1-\epsilon}{4\bar \chi}\,{\mathbf {H}}_{t,T}(\Theta_t). 
		\end{align}
		with probability at least $1/4$ for a number $\bar \chi$.  
	\end{theorem}
	
	The proof of this theorem and definition of $\bar \chi$ is rather involved and inspired by \cite{spielman2011graph}. The spectral bound \eqref{eq:spectralbound} can be used to obtain performance bounds for the estimation measures. 
	
	\begin{theorem} \label{thm:perf} Under the settings of Theorem \ref{theorem:sparse}, 
		the  estimation quality losses compared to the case where all features in $\Theta_t$ are used satisfy  
		\begin{eqnarray}
		\dfrac{\rho_v(\mathbf H_{t,T}(\Theta_t))-\rho_v(\mathbf H_{t,T}(\Phi_t))}{\rho_v(\mathbf H_{t,T}(\Phi_t))}  & \leq &       {\dfrac{4\Sp\bar{\chi}}{1-\epsilon}} -1 \label{eq:comega_bound_1} \\
		{\rho_e(\mathbf H_{t,T}(\Theta_t))-\rho_e(\mathbf H_{t,T}(\Phi_t))}  &\leq&    n \log \left ( \dfrac{4\Sp \bar{\chi}}{1-\epsilon} \right ) \label{eq:comega_bound_rhoe_1}  \\
		\dfrac{\rho_{\lambda}(\mathbf H_{t,T}(\Theta_t))-\rho_{\lambda}(\mathbf H_{t,T}(\Phi_t))}{\rho_{\lambda}(\mathbf H_{t,T}(\Phi_t))}  &\leq &     {\dfrac{4\Sp\bar{\chi}}{1-\epsilon}}-1, \label{eq:comega_bound_33} 
		\end{eqnarray}   
		with probability at least $1/4$. 
	\end{theorem}

	\subsection{Implementation of Algorithm}
	
	{The nature of the performance guarantees provided in Theorem \ref{theorem:sparse} and \ref{thm:perf}   motivates us to run the algorithm with multiple random seeds, i.e., by conducting Monte-Carlo simulations. To this end, we choose a design that corresponds to the minimal value of the estimation measure of interest.  We inspect that there are (at least) two steps during the feature selection process that are amenable to  parallel implementation: (i) evaluation of   sampling probabilities $\pi_f$ for different features can be done in parallel, and (ii) independent executions of Algorithm \ref{alg:execution} for the purpose of finding different designs can be conducted in parallel. }

	\subsection{Time-Complexity Analysis} \label{subsec:rt}
	
	To find the sampling probabilities, we need $O(N_tT^3)$ operations, where $N_t=|F_t|$ is the number of available features at time $t$. One execution of Algorithm \ref{alg:execution} requires $O(q T^2)$ operations. Evaluation of any of these estimation measures requires $O(T^3)$ operations. Therefore, if we run $p$ independent samples of this algorithm, we will need $O(pqT^2+pT^3)$ operations. Hence, the overall feature selection will require $O(N_tT^3+pqT^2+pT^3)$ operations. 
	
	For comparison purposes, we also analyze the time complexity of feature selection using the greedy method of \cite{carlone2018attention}. For this method, iteratively, we should examine all candidates and find the feature whose addition will enhance the estimation quality more than the remaining features. This method requires $O(qN_tT^3)$ operations. In the worst-case,  $q=O(N_t)$. Thus, in the worst case, its time complexity is $O(N_t^2 T^3)$, i.e., quadratic in the number of available features. This suggests that the random sampling using the leverage scores can potentially be faster than the greedy method (see next section for a numerical example).

	\section{Test Case} \label{sec:example}
	
	We bring the details of a numerical experiment, which is conducted to demonstrate the effectiveness of the feature-selection approach via Algorithm \ref{alg:execution}. 
	
	\begin{figure}[t]
		\begin{center}
			\includegraphics[width=8.2cm]{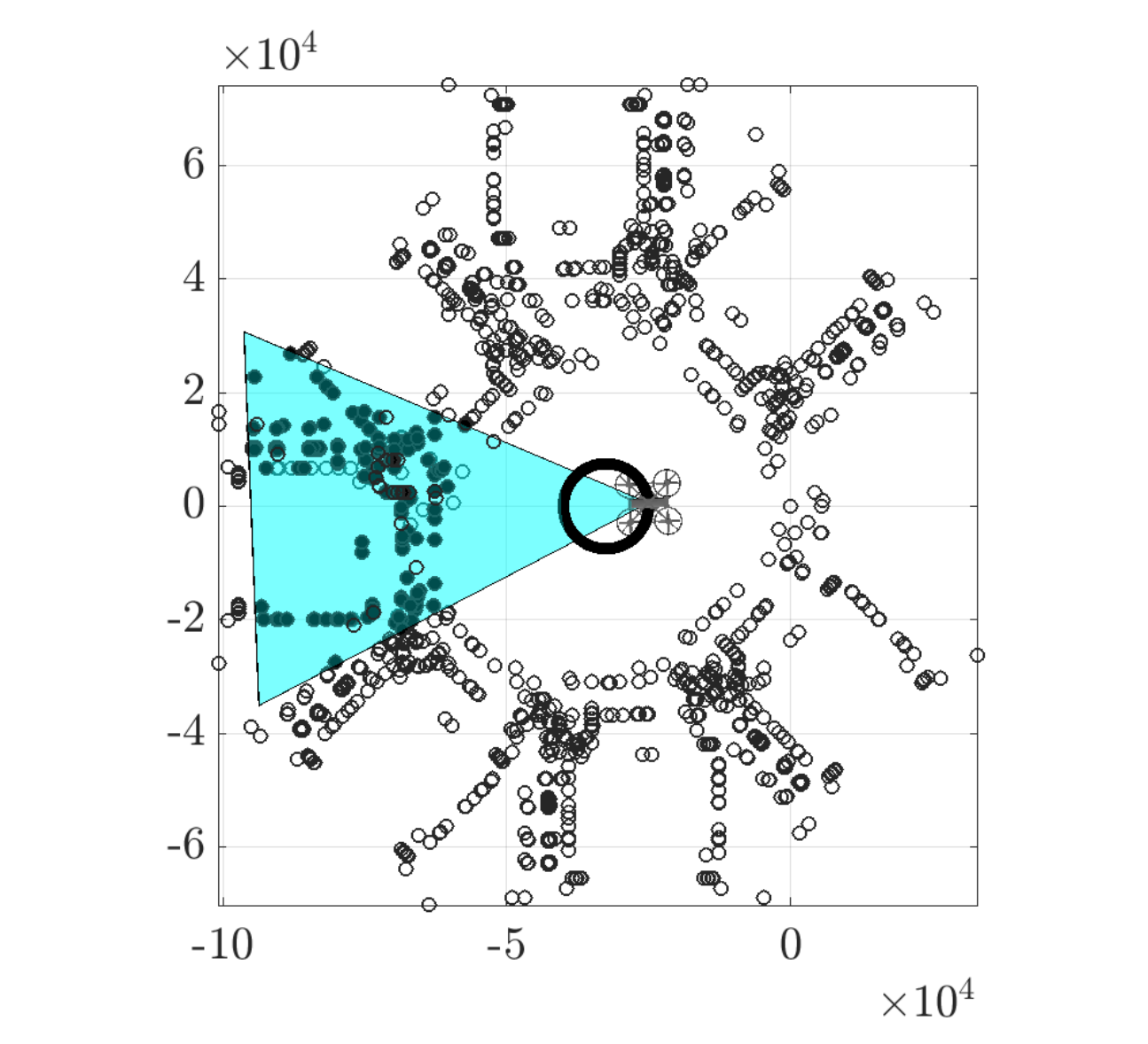}
			\caption{ The top view of the navigation environment. Note that the 3D reference curve is seen as a circle from this view. } 
			\label{fig:envup}
		\end{center}
	\end{figure}
	
	\begin{figure}[t]
		\begin{center}
			\includegraphics[width=8.1cm]{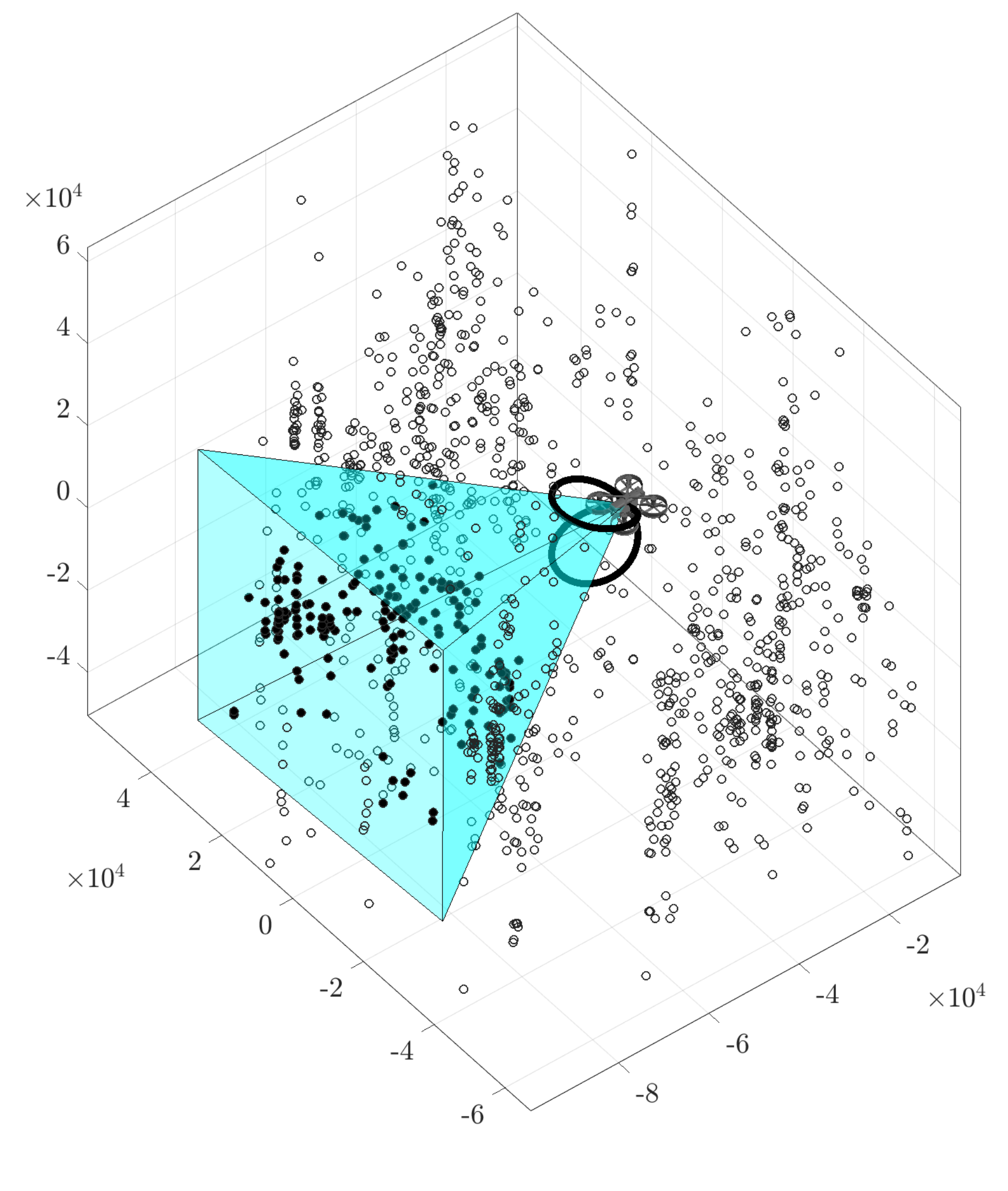}
			\caption{A snapshot of the environment.  The blue pyramid demonstrates the camera's field of view. The features that are inside the frame at this time are highlighted. The deformed $8$-shaped curve is the reference path as parametrized in \eqref{eq:refpath} (see Fig. \ref{fig:envup} for the top view as well). The robot is also rotating according to \eqref{eq:rot}. }        \label{fig:figenv}
		\end{center}
	\end{figure} 
	
	\begin{figure*}[t]
		\begin{center}
			\includegraphics[width=12cm]{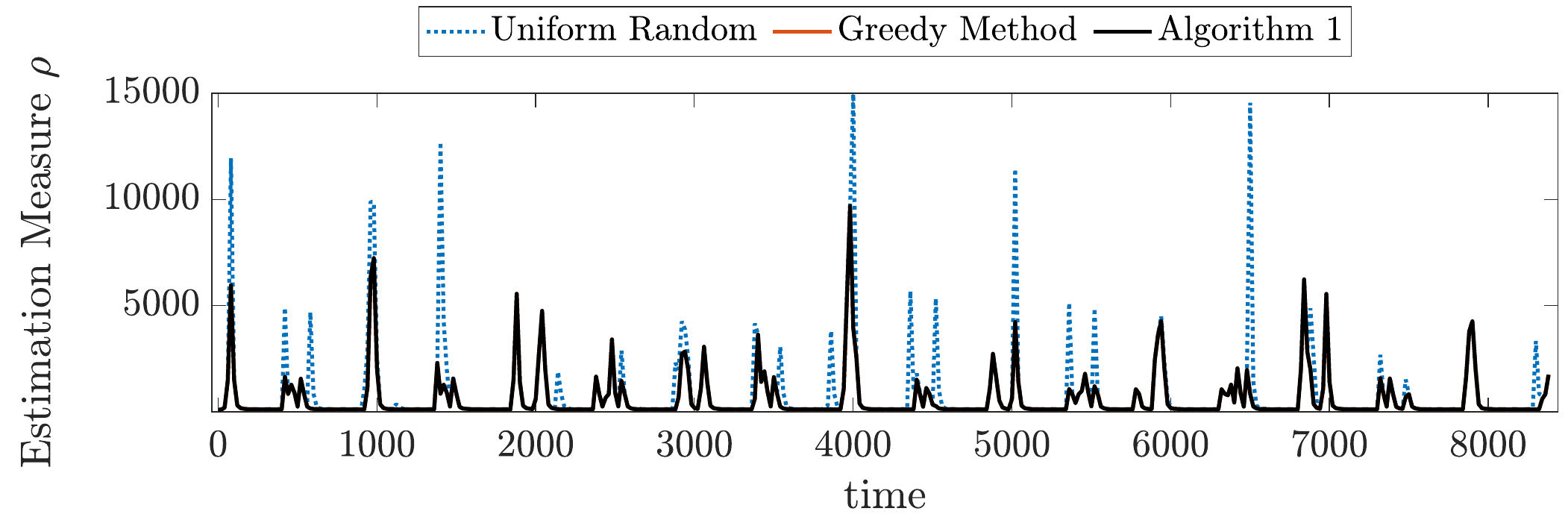}
			\caption{The estimation measure values resulting from three method. The curves corresponding to the  to the greedy method and the proposed method may not be distinguished in this plot. } 
			\label{fig:figonenu}
		\end{center}
	\end{figure*} 
	
	\begin{figure}[t]
		\begin{center}
			\includegraphics[width=8cm]{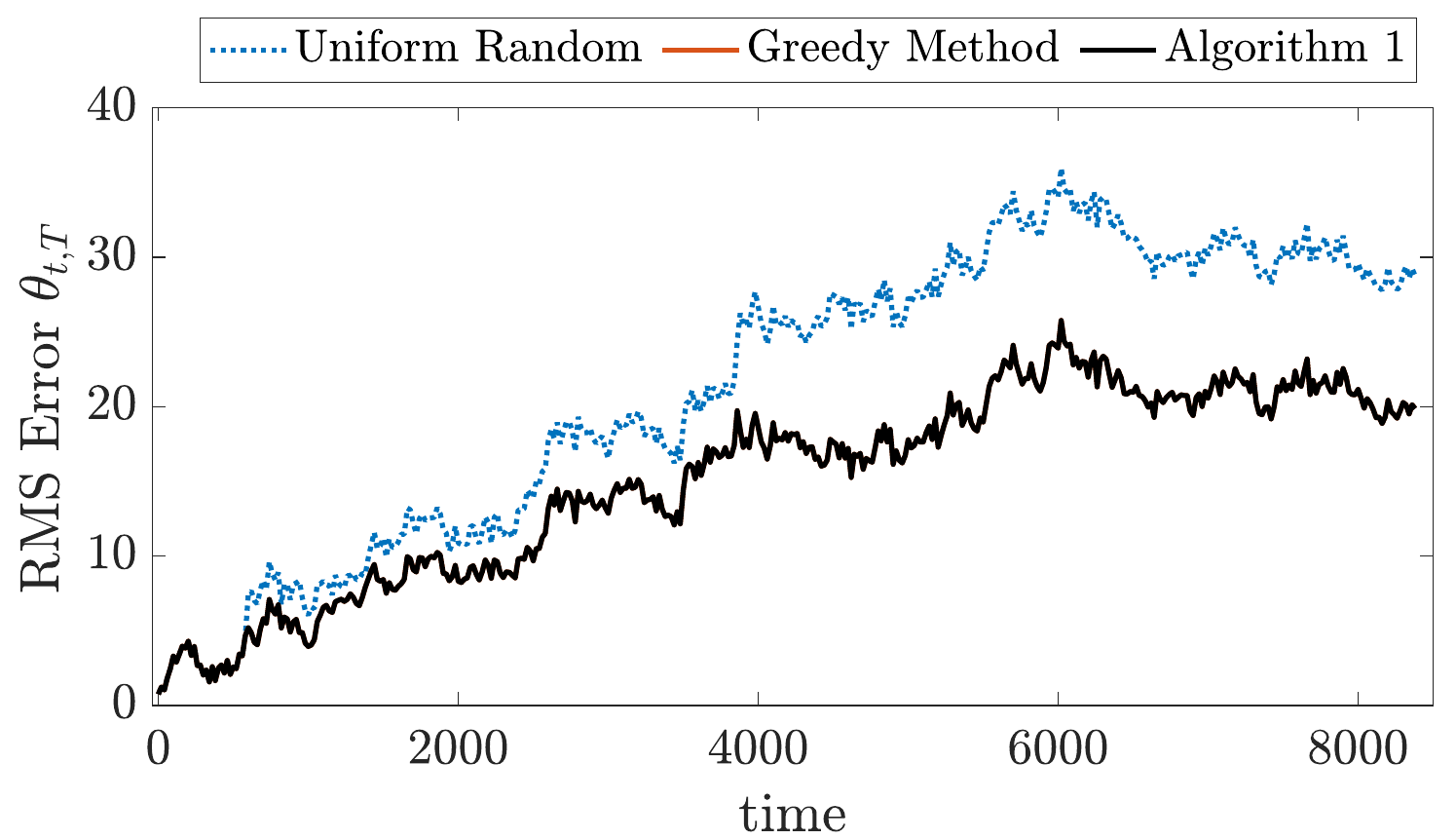}
			\caption{The  RMS error in the positions of the robot resulting from feature selection using different methods. The curces corresponding to the greedy method and Algorithm 1 may not be distinguished in this plot.   }
			\label{fig:rmsevst}
		\end{center}
	\end{figure}
	
	\subsection{Model and Environment Description}
	
	We consider a robot that is translating and rotating. Let us denote its position vector by $x_\tau^T=[p_\tau,y_\tau,z_\tau]$\footnote{We do use the letter $p$ for the first coordinate to prevent conflict with use of position vector $x$. }. We suppose that the high-level   dynamics of robot follow 
	\begin{align}
	\left \{\begin{array}{l}
	p_{\tau+1}= p_\tau+u_{\tau}^p+\delta_\tau^p \\
	y_{\tau+1}= y_\tau+u_{\tau}^y+\delta_\tau^y \\
	z_{\tau+1}= z_\tau+u_{\tau}^z+\delta_\tau^z
	\end{array} \right .,
	\end{align}
	where $  u_\tau^T:=[u_{\tau}^p,u_{\tau}^y,u_{\tau}^z]$ is the input command signal and ${\delta}_\tau^T:=[\delta_{\tau}^p,\delta_\tau^y,\delta_\tau^z]$ is the random process describing the uncertainty propagation due to the motion of the robot (compare to \eqref{eq:robotdynamics}). We consider the robot that is planning to move in the following reference path
	\begin{align} \label{eq:refpath}
	\left \{\begin{array}{l}
	p_{\tau}^{\mathrm{ref}}= p_0+R \cos(\omega \tau) \\
	y_{\tau}^{\mathrm{ref}}= R \sin(\omega \tau) \\
	z_{\tau}^{\mathrm{ref}}= R \sin(\omega \tau/2) 
	\end{array} \right ., 
	\end{align}
	which looks like a deformed $3$-D  number $8$ (see Fig. \ref{fig:figenv}). 
	To set the control inputs, we  set 
	\begin{align}
	\left \{\begin{array}{l}
	u_\tau^p=p_{\tau+1}^{\mathrm{ref}}-\bar \mu_\tau^p \vspace{1mm} \\
	u_\tau^y=y_{\tau+1}^{\mathrm{ref}}-\bar \mu_\tau^y \vspace{1mm} \\
	u_\tau^z=z_{\tau+1}^{\mathrm{ref}}-\bar \mu_\tau^z  
	\end{array} \right .. 
	\end{align}
	Moreover, we suppose that the Euler angles describing the absolute orientation of the   camera at time $\tau$ are given by 
	\begin{align}\label{eq:rot}
	\left \{\begin{array}{l}
	\alpha_\tau = {2\pi}  \sin \left ( \omega_r \tau \right ) \vspace{1mm}
	\\ 
	\beta_\tau =-\dfrac{\pi}{2}+\dfrac{\pi}{20} \sin \left ( \omega_r \tau  \right ) \\
	\gamma_\tau =0
	\end{array} \right .,
	\end{align}
	where the sequence of rotations is $z$-$y$-$p$. The visible landmarks in the environments consists of $1752$ points in the space, which is constructed by by putting a circular array of randomly sampled points in a $3$-D model of a room\footnote {the \textrm{STL} graphical file is adapted from \url{https://grabcad.com/library/room-blender-test-1}}. Two views from a snapshot of the environment have been illustrated in Fig. \ref{fig:envup} and Fig. \ref{fig:figenv}.  We set parameters $R=7500$,   $\omega=0.08$, $\omega_r=0.0064$, $\sigma=0.1$ and $\Lambda_t=\mathrm{diag}(4,4,16)$ and initial covariance to be $\Sigma_{0}=I$.


	\subsection{Different Approaches for Feature Selection}
	
	We consider the navigation setup for $400$  time horizons each of length $T=20$. The overall simulation consists of doing almost $107$ full turns around the $3$-D path of interest. 
	For each horizon, after finding the eligible features to track (i.e., features with full rank information matrix $(\mathbf E_{t,T}^f)^T   \mathbf E_{t,T}^f$), we select at most half of the features. For each horizon, we do this task via three different methods:
	
	\noindent {\it (i)   randomized sampling by leverage scores:} we run Algorithm \ref{alg:execution} for $p=50$ independent experiments and choose the set $\Phi_t$ which induces the minimal value of the estimation measure. 
	We denote the CPU time spent on this task by $\tau_t$.  
	
	\noindent {\it (ii)   randomized sampling using uniform probabilities:} we run Algorithm \ref{alg:execution} for $p=50$ independent experiments, except that instead of evaluating the probabilities, we assume an equal probability for each feature to be sampled. Similar to the previous case, we choose the design that induces minimal value of the estimation measure. We denote this design by $\Phi_t^{u}$, where $u$ stands for uniform probabilities. 
	Similarly, we denote the corresponding CPU time for this task by $\tau_t^u$. 
	
	\noindent {\it (ii)   greedy method:} the features are added one-by-one, where at each iteration the feature which enhances the estimation quality the most is selected \cite{carlone2018attention}. This method produces a single design for the feature selection denoted by $\Phi_t^g$, where $g$ stands for the greedy method. 
	
	In this example, we consider the estimation measure $\rho_v$ as the monotone function governing the feature selection.

	\subsection{Metrics for Comparison of Methods}
	
	Many researchers have observed that the greedy method over-performs other approaches in several similar combinatorial problems \cite{carlone2018attention,lin2014algorithms}. In general,  the brute-force method in these settings is computationally infeasible\footnote{For instance, for a choice of 25 features out of 50 candidates we have to examine more than $10^{14}$ possible combinations. In fact, finding the brute-force solution becomes rapidly computationally prohibitive as the size of the candidate set grows.}, we select the greedy method as the base approach. 
	Moreover, to have a clear understanding of the error in the position, we define the root mean squared error (RMSE) as we define 
	\begin{align}
	\theta_{t,T}&:=\dfrac{1}{3(T+1)}\sqrt{ \sum_{\tau=t}^{t+T} \|x_\tau- \mu_t \|_2^2},
	\end{align} 
	for $t \in \{0,T,2T,\dots\}$. 
	We define similar error indices for the uniform random (method {\it (ii)}) and the greedy method  (method {\it (iii)}) as well and denote them by ${\theta}_{t,T}^u$ and ${\theta}_{t,T}^g$, respectively. To compare the relative difference of these values, we use 
	\begin{align}
	\phi_{t,T}&:=\dfrac{\theta_{t,T}-  {\theta}_{t,T}^g  } {{\theta}_{t,T}^g} \times 100.
	\end{align}
	Similarly, we define $\phi_{t,T}^u$, which compares the value of RMSE   resulting from  the totally random choice of feature  with   greedy selection. Finally, to compare the CPU times, we look at  
	\begin{align}
	\kappa_t&:=\dfrac{\tau_t}{ \tau_t^g},  
	\end{align} 
	which represent the ratio of the CPU time spent in    methods {\it (i)}   to the one spent by the greedy method. Similarly, we use $ {\kappa_t^u}$  to   compare the time spent by method {\it (ii)} with method {\it (iii)}.

	\subsection{Numerical Results}
	

	In Fig \ref{fig:figonenu}, we show the resulting values of the estimation measure versus time, which demonstrate that the estimation measure resulting from Algorithm \ref{alg:execution} is almost identical to the estimation measure resulting from the greedy method. This is not the case for the totally random choice of features, in which larger spikes can be observed.

\begin{figure}[t]
		\begin{center}
			\includegraphics[width=8.2cm]{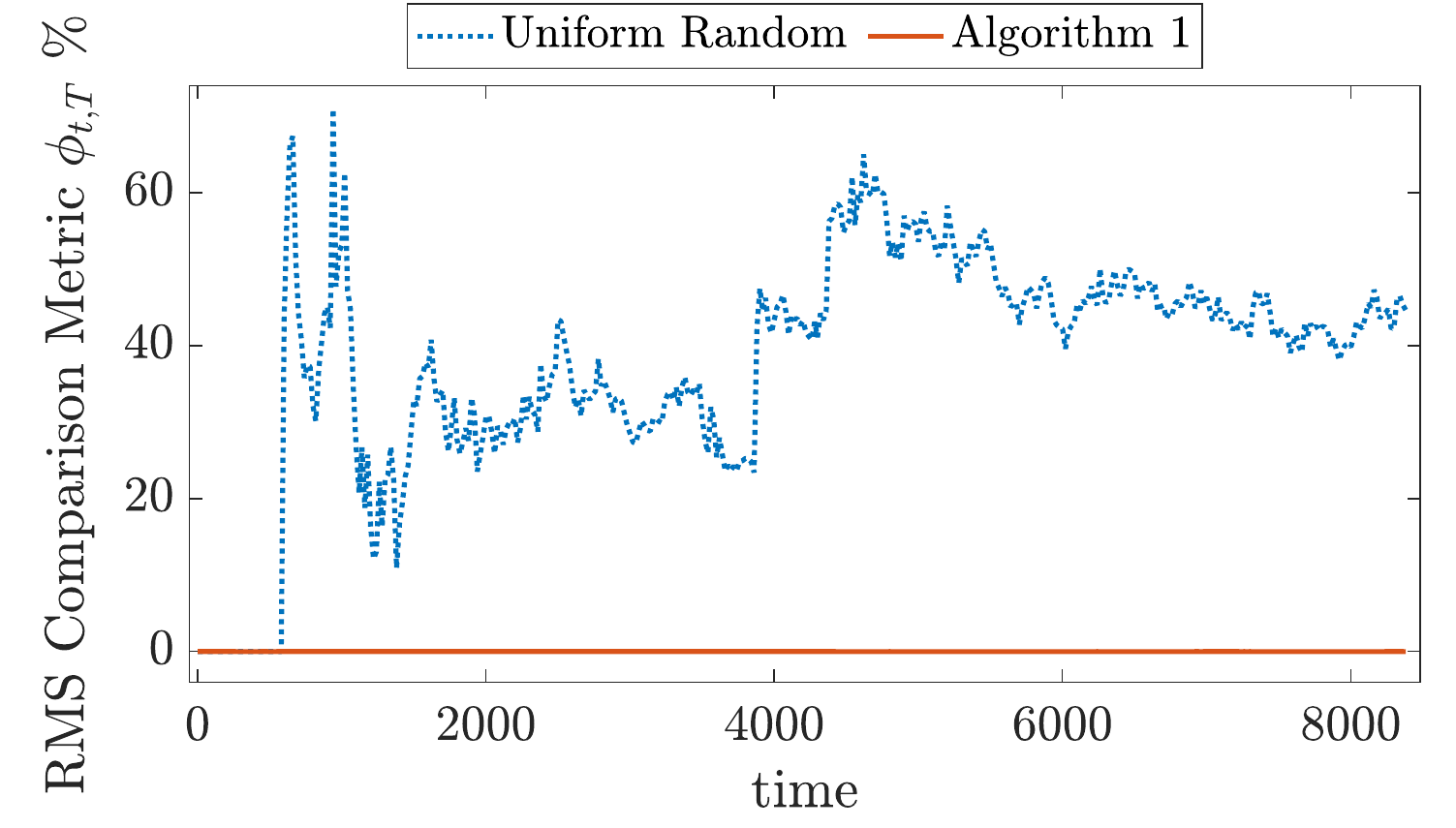}
			\caption{Comparison of the RMS error resulting from the uniform random method and our proposed approach with the greedy method.  This plot shows a speed-up by almost an order of magnitude in the feature selection using Algorithm \ref{alg:execution}.} 
			\label{fig:figphi}
		\end{center}
	\end{figure}

	In Fig. \ref{fig:rmsevst}, we illustrate the values of the RMS error versus times for these methods. The errors corresponding to the feature selection using Algorithm \ref{alg:execution} is    very close to those values for the greedy method, while the totally random method may result in larger errors. 
	This shows that in this example, not only the choice of features is a non-trivial computational task, but also our algorithm functions with quality and reliability that is very close to these factors in the case of the greedy method. The   metric $\phi_{t,T}$ is also illustrated in Fig. \ref{fig:figphi}, which quantifies these deviations.

	\begin{figure}[t]
		\begin{center}
			\includegraphics[width=8.2cm]{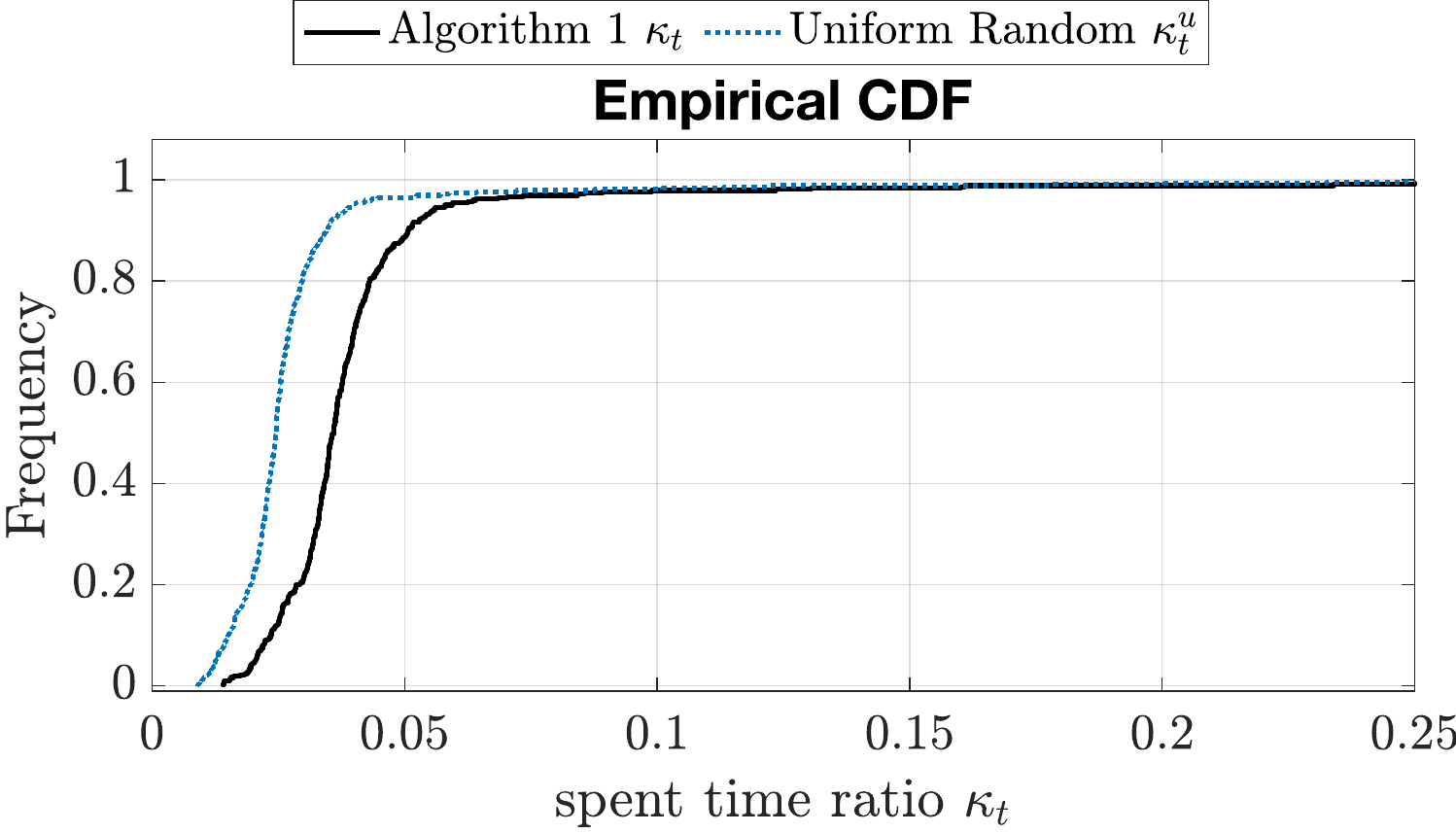}
			\caption{The empirical CDF's for parameters $\kappa_t$ and $\kappa_t^u$, which show the ratio of the spent CPU time for the random sampling (including all independent 50 experiments) to the greedy method. The plot show that in this example the random sampling is faster than the greedy method by almost an order of magnitude.    }
			\label{fig:figkappa}
		\end{center}
	\end{figure}
	
	Finally, we compare the CPU times spent on each method.  In Fig. \ref{fig:figkappa}, we demonstrate the experimental CDF's of the parameters $\kappa_t$ and $\kappa_t^u$, which show that the randomized methods are considerably faster than the greedy method in most cases. For instance, these data suggest that Algorithm \ref{alg:execution}   has been more than $20$ times faster than the greedy method in about $85\%$ of the assigned tasks, while in most cases it has been least $10$ times faster. Note that the time  includes  running the random sampling algorithms for $50$ independent experiments

	

	\section{Conclusion and Discussion}  \label{sec:conc}
	
	We propose a randomized algorithm for visual feature selection over a fixed-length moving time horizon. The idea is to associate a sampling  probability to each candidate feature,  randomly sample a subset of features according to these probabilities for a number of independent experiments, and select the outcome with the best estimation quality.  We would like to discuss a few remarks.
	
	If the estimation measure enjoys submodularity, then the greedy method provides a performance guarantee compared to the optimal solution \cite{carlone2018attention,tzoumas2016sensor}. However, it is known that certain measures, e.g., $\rho_v$, are not submodular \cite{olshevsky2018non}. Moreover, in our work, we offer a different type of  performance guarantee.  Theorem \ref{thm:perf}   compares the estimation quality to the case that we leverage all features for tracking. Nevertheless, our extensive numerical simulations assert that the resulting estimation quality from our algorithm and that of the greedy method are often close to each (see Fig. \ref{fig:figonenu} for example). Further research is required to uncover the practical and theoretical differences of this randomized algorithm and the greedy method. 
	
	
	Based on the time-complexity analysis provided in Section \ref{subsec:rt}, if the number of selected features is small, the computational requirements of our randomized sampling method will be comparable to greedy methods. However, in the worst case, the greedy-method scales quadratically with the number of candidate features (i.e., scaling with $N_t^2=|\Theta_t|^2$). This justifies the significant speed-up in the feature-selection that was observed in our numerical simulations (see Fig. \ref{fig:figkappa}). The low time-complexity of our method  opens up new opportunities for real-time implementation of this algorithm and utilizing it for agile robot navigation.

	    \section*{Appendix}
	    
	    \subsection*{Appendix A: Proof of Lemma \ref{lem:one}}

	    After one episode of motion in the horizon, one can verify that the updates to covariance and the mean are given by 
	    \begin{align} \label{eq:sigmaupdate}
	    \bar { \Sigma}_{t+1} &\,=\,  { A}_{t+1}  \bar { \Sigma}_{t}    { A}_{t+1}^T +  {\Lambda}_{t+1}   , \\
	    \bar {\mu}_{t+1}&\, =\,  \bar { \Delta}_{t+1},    \label{eq:qbarupdateap}
	    \end{align}
	    respectively. Because these update laws also work   for any time instant in the horizon, we can use the similar update 
	    \begin{align}  \label{eq:sigmaupdate_tauap}
	    \bar { \Sigma}_{\tau} &\,=\,  { A}_\tau  \bar { \Sigma}_{\tau-1}    { A}_\tau^T +   {   \Lambda}_\tau   , \\
	    {\mu}_\tau&\, =\,  \bar  { \Delta}_\tau,     
	    \end{align}
	    %
	    %
	    for every instant $\tau \in \{t+1,\dots,t+T\}$. 
	    To fully identify   covariance matrix $ \bar {\mathbf \Sigma}_{t,T}$, we can show that for all $\tau_1,\tau_2 \in \{t,t+1,\dots,t+T\}$, with $\tau_1<\tau_2$, 
	    \begin{align}\label{eq:offdiag}
	    &\mathbb{E} \{  x_{\tau_1} x_{\tau_2}^T \}-\mathbb{E} \{ x_{\tau_1}  \}\mathbb{E} \{ x_{\tau_2}^T \}=\\ &\left ( \prod_{i=1}^{\tau_2-\tau_1} { A}_{\tau_2-i-1} -  I \right)   \bar { \Sigma}_{\tau_1} \notag :=  \bar { \Sigma}_{\tau_1,\tau_2}. 
	    \end{align}
	    The proof of this relationship can be conducted by induction. 
	    Then, we observe that the covariance matrix has the structure 
	    \begin{align} 
	    \bar {\mathbf \Sigma}_{t,T}=
	    \begin{bmatrix} 
	    \bar { \Sigma}_t & \bar { \Sigma}_{t,t+1} &\dots  & \bar { \Sigma}_{t,t+T} \\
	    \bar { \Sigma}_{t,t+1}^T  & \bar { \Sigma}_{\tau+1} & \dots  &\bar { \Sigma}_{t+1,t+T}  \\
	    \vdots & \vdots & \ddots & \vdots \\
	    \bar { \Sigma}_{t,t+T}^T & \bar { \Sigma}_{t+1,t+T}^T & \dots & \bar { \Sigma}_{t,T}
	    \end{bmatrix},
	    \end{align}
	    where the diagonal and  off-diagonal matrix terms are supposed to be evaluated by iterative application of $\eqref{eq:sigmaupdate_tauap}$ and \eqref{eq:offdiag}, respectively.  
	    For the mean, the update is simply achieved by stacking the updated mean variables in \eqref{eq:qbarupdateap}. This completes the proof.

	    \subsection*{Appendix B: Proof of Lemma \ref{lem:updateb}}
	    
	    The proof of this lemma is based on an approach similar to one given in \cite{carlone2018attention}. Note that initial  information matrix of the stacked variable $[\mathbf x_{t,T} ^T~  y_f^T]^T$ is
	    $$
	    \begin{bmatrix}
	    \bar {\mathbf H}_{t,T} & 0 \\
	    0 & 0
	    \end{bmatrix},
	    $$
	    as we assume no information regarding the location of the feature $f$\footnote{ Informally, this is due to the fact that we are not building a map out of these observations. Instead, we only track them to enhance the quality of our estimation from the position of the robot. }. Then, the update to the information matrix based on the observation model is of the form 
	    \begin{align*}\widetilde {\mathbf H}_{t,T}:=
	    \begin{bmatrix}
	    \bar {\mathbf H}_{t,T} & 0 \\
	    0 & 0
	    \end{bmatrix} +
	    \begin{bmatrix}
	    (\mathbf F_{t,T}^f)^T \mathbf F_{t,T}^f & (\mathbf F_{t,T}^f)^T \mathbf E_{t,T}^f \\
	    (\mathbf E_{t,T}^f)^T \mathbf F_{t,T}^f & (\mathbf E_{t,T}^f)^T \mathbf E_{t,T}^f
	    \end{bmatrix},  
	    \end{align*} 
	    (e.g. see \cite{thrun2004simultaneous} to see its derivation based on the Bayes law). Given the structure of this matrix, using the Schur complement, one can show that 
	    \begin{align*}
	    (\tilde {\mathbf H}_{t,T})^{-1}=\begin{bmatrix}
	    \mathbf H_{t,T}^{-1} & -\mathbf H_{t,T}^{-1} \mathbf C_{t,T}^f  \\
	    \star_1 & \star_2
	    \end{bmatrix},
	    \end{align*}
	    where the value of the starred matrices will not be needed and we have 
	    \begin{align}
	    & \mathbf H_{t,T}=\bar {\mathbf H}_{t,T}+\mathbf H_{t,T}^f  \\ 
	    & \mathbf C_{t,T}^f:=(\mathbf F_{t,T}^f)^T \mathbf E_{t,T}^f \big ( (\mathbf E_{t,T}^f)^T \mathbf E_{t,T}^f \big )^{-1}. 
	    \end{align}
	    Now, note that the update to the information vector of the stacked variable $[\mathbf x_{t,T} ^T~  y_f^T]^T$ is of the form 
	    \begin{align*}\widetilde {\mathbf b}_{t,T}^T:=
	    \begin{bmatrix}
	    \bar {\mathbf b}_{t,T}^T  \\
	    0 
	    \end{bmatrix} +
	    \begin{bmatrix}
	    (\mathbf F_{t,T}^f)^T \mathbf z_{t,T}^f   \\
	    (\mathbf E_{t,T}^f)^T \mathbf  z_{t,T}^f
	    \end{bmatrix}.  
	    \end{align*} 
	    Therefore, the update for the mean of the stacked variable is given by 
	    \begin{align*}
	    \widetilde {\boldsymbol \mu}_{t,T}&=(\tilde {\mathbf H}_{t,T})^{-1}  \widetilde {\mathbf b}_{t,T}^T \\&=\begin{bmatrix}
	    \mathbf H_{t,T}^{-1} \big ( \bar{\mathbf b}_{t,T}^T+ (\mathbf F_{t,T}^f)^T -\mathbf C_{t,T}^f (\mathbf E_{t,T}^f)^T  \big )\mathbf z_{t,T}^f) \\
	    \star_3
	    \end{bmatrix} \\
	    & =\begin{bmatrix}
	    \mathbf H_{t,T}^{-1} (\bar {\mathbf b}_{t,T}^T+  \mathbf B_{t,T}^f  \mathbf z_{t,T}^f)  \\
	    \star_3 
	    \end{bmatrix}, 
	    \end{align*}
	    where in the last one we have used the definition of $\mathbf B_{t,T}^f$ and the starred element will not be used.  
	    Based on the partition of the stacked variable, we inspect that 
	    \begin{align}
	    \boldsymbol{\mu}_{t,T}=\mathbf H_{t,T}^{-1} (\bar {\mathbf b}_{t,T}+  \mathbf B_{t,T}^f  \mathbf z_{t,T}^f).
	    \end{align}
	    Hence, based on the definition of the information vector, by inspection, we get that 
	    \begin{align}
	    \mathbf b_{t,T}^T= \bar {\mathbf b}_{t,T}^T+  \mathbf B_{t,T}^f  \mathbf z_{t,T}^f.
	    \end{align}
	    This completes the proof of the lemma.

	    \subsection*{Appendix C: Proof of Theorem \ref{theorem:sparse}}
	    
	    For all features $f \in \Theta_t$, the corresponding matrix $\bar {\mathbf H}_{t,T}^f$ can be decomposed as
	    \begin{align}\label{eq:dec}
	    \bar {\mathbf H}_{t,T}^f=\sum_{i=1}^{n} \mathbf h_{t,T}^{f,i} (\mathbf h_{t,T}^{f,i})^T :=\sum_{i=1}^{n}   \bar {\mathbf H}_{t,T}^{f,i} ,
	    \end{align}
	    where   vectors $\mathbf h_{t,T}^{f,i}  \in \R^n$ are orthogonal to each other for each $\bar {\mathbf H}_{t,T}^f$. Moreover, we define the refined leverage scores as follows. 
	    \begin{definition} For a given set of features $\Theta_t$ and decomposition of information matrices given in \eqref{eq:dec}, the refined leverage scores are  nonnegative numbers defined by 
	        \begin{align}
	        r_{fi}:= \mathrm{Tr}\left ( {\mathbf {H}}_{t,T}(\Theta_t)^{-1} \bar {\mathbf H}_{t,T}^{f,i} \right ).
	        \end{align}
	        for every $f \in   \Theta_t$ and $i \in \{1,\dots,n\}$. 
	    \end{definition}
	    
	    \begin{algorithm}[t]
	        \caption{Randomized Feature Selection \\ ~~~~~(Analysis Version with Additional Lines)}
	        \label{alg:sparsification}
	        \begin{algorithmic}  
	            \algnewcommand\algorithmicinitz{\textbf{initialize:}}
	            \algnewcommand\Init{\item[\algorithmicinitz]}
	            \renewcommand{\algorithmicrequire}{\textbf{input:}}
	            \renewcommand{\algorithmicensure}{\textbf{output:}}
	            \Require initial information matrix $\bar {\mathbf H}_{t,T}$ \\ ~~~~~~set of available features $\Theta_t$, number of samples $q$
	            \Ensure  selected features $\Phi_t$,   information matrix $\mathbf H_{t,T}$  \\        
	            ~~~~~~~~set of indices $\hat \Phi_t$,   weight function $w_t$ \vspace{0.15cm} \Comment 
	            \Init $\hat \Phi_t=\varnothing$,  $\mathbf H_{t,T}=\bar{\mathbf H}_{t,T}$ \\
	            ~~~~~~~~~~$w_t(.,.) = 0$ \Comment 
	            
	            \For {$k=1$ to $q$}
	            \State sample a feature from $\Theta_t$ using distribution $\pi$~$\rightarrow f$
	            \State select the corresponding update  matrix 
	            $$
	            \mathbf H \leftarrow   {\mathbf H} _{t,T}^f 
	            $$
	            
	            \State sample from $1$ to $n$   with distribution $p_{f}$ $\rightarrow i$ \Comment 
	            \State update weight function:   \Comment 
	            \State ~~~~~~~~~~~~~~   $
	            w_t(f,i)  \leftarrow w_t(f,i)  + (q\pi_{fi})^{-1}$  \vspace{2mm}  \Comment  
	            \If {$f \notin \Phi_t$,} 
	            \State add $f$ to $\Phi_t$  
	            \State update the information matrix:
	            $$ \mathbf H_{t,T}  \leftarrow \mathbf H_{t,T} +\mathbf H $$
	            \EndIf
	            \vspace{0.05cm}
	            \If {$(f,i) \notin \hat \Phi_t$,} \Comment 
	            \State add $(f,i)$ to $\hat \Phi_t$  
	            \vspace{0.05cm} \Comment 
	            \EndIf \Comment
	            \EndFor
	            
	        \end{algorithmic}
	    \end{algorithm}  
	    
	    Similarly, let us denote the refined sampling probabilities by
	    $
	    \pi_{fi}:={r_{fi}}/{n}.
	    $
	    Due to the orthogonality of the vectors $\mathbf h_{t,T}^{f,i}$, one inspects that
	    \begin{align}\label{eq:pr_sum}
	    r_f=\sum_{i=1}^{n} r_{fi},~~\pi_f=\sum_{i=1}^{n} \pi_{fi},
	    \end{align}
	    for $f\in   \Theta_t$. We define 
	    $$
	    p_f(i)=p_{fi}=\dfrac{\pi_{fi}}{\displaystyle \sum_{i=1}^n \pi_{fi}}=\dfrac{\pi_{fi}}{\pi_f}.
	    $$
	    We observe that over each $f\in   \Theta_t$,   numbers $p_{fi}$ also constitute a probability distribution (i.e, they sum up to $1$). Now, we modify Algorithm \ref{alg:execution} to get Algorithm \ref{alg:sparsification}, wherein lines labeled with symbol $\triangleright$  are   only required    for performance analysis and do not need to be conducted during the execution of the algorithm; i.e., removing those lines from Algorithm \ref{alg:sparsification} will give us Algorithm \ref{alg:execution}. Algorithm \ref{alg:sparsification} assigns values to a weight function that lets us assess the quality of estimation.
	    The  weight function $w_t: ~\hat \Phi_t \rightarrow \R_{+}$ is a   bounded random variable, where  $w_t(f,i)$
	    may take different realizations drawn from  $\big\{ p (q \pi_{fi})^{-1}~\big|~p=0,1,\dots,q \big\}$.
	    
	    Now, we build up our proof based on  some of the steps taken in the proof of Theorem 5 in \cite{batson2013spectral} and  steps from  \cite{mousavi2018space}, wherein the authors prove a similar result for the case that selected matrices are rank-one.  Using the leverage scores  \eqref {resistent.def}, one can verify that because $\bar {\mathbf H}_{t,T}^{f,i}$ for each $(f,i) \in \hat \Phi_t$ is rank-one, then 
	    by applying similar steps to the proof of Theorem 5 in \cite{batson2013spectral},  with probability at least $1/2$ we have
	    \begin{align}\label{eq:spec}
	    (1-\epsilon)\mathbf H_{t,T}(\Theta_t) \Sp \preceq \Sp \Omg_w  ,
	    \end{align}
	    where $\Omg_w$ is given by 
	    \begin{align}
	    \Omg_w \Sp =& \Sp  \sum_{(f,i) \in \hat \Phi_t}\Sp w_t(f,i) \Sp \bar {\mathbf H}_{t,T}^{f,i} \Sp = \Sp \sum_{(f,i) \in \hat \Phi_t} \Sp \frac{\phi_{fi}}{q\pi(f )} \Sp \Sp \bar {\mathbf H}_{t,T}^{f,i},
	    \end{align}
	    and $\phi_{fi}\geq 0$ is the frequency of the times that feature $f$ and then index $i \in \{1,\dots,n\}$ are sampled by Algorithm \ref{alg:sparsification}. 
	    %
	    Now, consider the matrices
	    $$
	    {T} :=\begin{bmatrix}
	    (\mathbf h_{t,T}^{f,i})^T
	    \end{bmatrix}_{(f,i) \in \hat \Phi_t}, ~~{  W} =\mathrm{diag}\big(w_t(f,i)\big)\big|_{(f,i)\in \hat \Phi_t}.
	    $$
	    Let us introduce an \textit{artificial} linear parameter estimation problem based on the model 
	    \begin{align}\label{eq:noisy_model2}
	    y=  T    \theta+ {\eta}
	    \end{align}
	    where observation is given by $  y $ and 
	    $ {\eta}$ is a zero mean  Gaussian measurement noise of independent components and  covariance
	    $
	    \mathbb{E}  \left \{  {\eta}  {\eta}^T  \right \}=   I$.  In the next step, let us consider   estimators  $\hat {   \theta}$ and $\tilde {   \theta}$ that are given by
	    \begin{align}
	    \hat{  \theta}   = & \big (  T ^T    \Sp   T \big )^{-1} \Sp   T^T    {y}. \\ 
	    \tilde{  \theta}   = & \big (  T ^T  {  W}   \Sp   T\big )^{-1} \Sp   T ^T {  W}   {y}.
	    \end{align}
	    It is straightforward to verify that both of them are unbiased estimators for $  \theta$. Since covariance of noise is $
	    \mathbb{E}  \left \{  {\eta}  {\eta}^T  \right \}=     I
	    $, the unweighted least-squares estimator is the optimal estimator.  Consequently, by Gauss-Markov theorem 
	    \begin{align}\label{eq:gaussmarkov}
	    \mathbb{E}\left \{\tilde {  \theta} \tilde {  \theta}^T\right \} \succeq  \mathbb{E}\left \{\hat {  \theta} \hat {  \theta}^T \right \}.
	    \end{align}
	    Now, we explicitly write down the two sides of  \eqref{eq:gaussmarkov}. First observe that 
	    \begin{align}
	    \mathbb{E}\left \{\hat { \theta} \hat { \theta}^T \right \}= \left ( T  ^T    \Sp  T   \right )^{-1}=\left (\sum_{(f,i) \in \hat \Phi_t} \bar {\mathbf H}_{t,T}^{f,i} \right )^{-1} :=\hat \Omg_{t,T}^{-1}.
	    \end{align}
	    Because $\Omg_w= T ^T  W  T $, we can also write 
	    \begin{align}\label{eq:exptilde}
	    \mathbb{E}\left \{\tilde { \theta} \tilde { \theta}^T \right \} = 
	    \Omg_w ^{-1}  T  ^T   W ^2  T  \Omg_w^{-1}. 
	    \end{align}
	    We define $\chi$ to be  a random variable given by
	    \begin{align*}
	    \begin{adjustbox}{max width=245pt}
	    $
	    \chi=\inf \Bigg\{{\gamma>0} \Bigg|     \displaystyle \sum_{(f,i) \in \hat \Phi_t } w_t(f,i)  \big(\gamma - w_t(f,i) \big)\Sp \bar {\mathbf H}_{t,T}^{f,i}  \succeq {\bf 0} \Bigg\}.
	    $
	    \end{adjustbox} 
	    \end{align*}
	    Moreover, we set $\bar{\chi}:=\mathbb{E}\left\{ \chi\right\}$. 
	    Let us isolate the   term in the middle of \eqref{eq:exptilde}. Based on  the definition of random variable  $\chi$, one can write 
	    \begin{align}\label{eq:chiinequality}
	    T  ^T  W ^2  T  \preceq \chi \Omg_w  
	    \end{align}
	    One can show that for every three positive-definite matrices $ X_1, X_2, X_3$ with $ X_1 \succeq  X_2$, inequality
	    \begin{align}\label{eq:threematrices}
	    X_3  X_1  {X_3} \succeq  X_3  X_2  {X_3}
	    \end{align}
	    holds. Combining \eqref{eq:exptilde}, \eqref{eq:chiinequality}, and \eqref{eq:threematrices} we find that 
	    \begin{align}\label{eq:chiboundone}
	    \mathbb{E}\left \{\tilde { \theta} \tilde { \theta}^T \right \} \leq \Omg_w ^{-1} (\chi\Omg_w)  \Omg_w^{-1} =\chi   \Omg_w^{-1}.  
	    \end{align}
	    Based on Markov inequality,  it holds that 
	    \begin{align}\label{eq:someboundomega}
	    \chi \Sp \leq \Sp \dfrac{1}{1-\frac{3}{4}} \Sp \mathbb{E}\{\chi\} \Sp = \Sp 4\bar \chi
	    \end{align}
	    with probability at least $3/4$. From \eqref{eq:chiboundone} and \eqref{eq:someboundomega}, we get
	    \begin{align}
	    \mathbb{E}\left \{\tilde { \theta} \tilde { \theta}^T \right \}  \preceq 4\bar \chi \Omg_{w}. 
	    \end{align}
	    On the other hand, because \eqref{eq:spec} holds, with probability at least $1/2$
	    By taking inverse, we get \begin{align}\label{eq:somebound4}
	    \Sp\Omg_w     \Sp \preceq (1-\epsilon)  ^{-1} {\mathbf {H}}_{t,T}(\Theta_t)^{-1} ,
	    \end{align}
	    If the event described in \eqref{eq:someboundomega} is denoted by $\mathfrak{A}$ and the event described by \eqref{eq:somebound4} is denoted by $\mathfrak{B}$, then
	    $$
	    \mathbb{P}(\mathfrak{A} \cap \mathfrak{B}) = \mathbb{P}(\mathfrak{A})+\mathbb{P}(\mathfrak{B})-\mathbb{P}(\mathfrak{A} \cup \mathfrak{B}) \geq \dfrac{3}{4}+\dfrac{1}{2} -1=\dfrac{1}{4}.
	    $$
	    Therefore both \eqref{eq:someboundomega} and \eqref{eq:somebound4} hold with probability at least $1/4$; i.e, the inequality 
	    $$
	    \mathbb{E}\left \{\tilde { \theta} \tilde { \theta}^T \right \}  \preceq 4\bar \chi (1-\epsilon)  ^{-1} {\mathbf {H}}_{t,T}(\Theta_t)^{-1} ,
	    $$
	    holds with probability at least $1/4$. Because \eqref{eq:gaussmarkov} holds, with probability at least $1/4$
	    \begin{align}
	    \hat \Omg_{t,T}^{-1}    \preceq 4\bar \chi (1-\epsilon)  ^{-1} {\mathbf {H}}_{t,T}(\Theta_t)^{-1}. 
	    \end{align}
	    Or equivalently, 
	    \begin{align}\label{eq:omghatbound}
	    \hat \Omg_t     \succeq  \dfrac{1-\epsilon}{4\bar \chi}  {\mathbf H}_{t,T}(\Theta_t) . 
	    \end{align}
	    As the final step, we observe that 
	    \begin{align}\label{eq:omgSineq}
	    \Omg_{t,T}(\Phi_t)&=\bar \Omg_{t,T}+ \sum_{f \in \Phi_t}   \mathbf{H}_{t,T}^f   \\ \notag
	    &\succeq \dfrac{|\Phi_t|}{N_t}  \bar \Omg_{t,T}+ \sum_{f \in \Phi_t}   \mathbf{H}_{t,T}^f \\ \notag
	    &=     \sum_{f \in \Phi_t}   \dfrac{1}{N_t}\bar \Omg_{t,T}+ \mathbf{H}_{t,T}^f=\sum_{f \in \Phi_t}\bar {\mathbf H}_{t,T}^f 
	    \\ \notag
	    &\succeq    \sum_{(f,i) \in \hat \Phi_t}    \bar {\mathbf H}_{t,T}^{f,i}=\hat \Omg_t .     
	    \end{align}
	    Combining \eqref{eq:omghatbound} and \eqref{eq:omgSineq} we conclude that 
	    \begin{align}
	    \Omg_{t,T}(\Phi_t)  \succeq \dfrac{1-\epsilon}{4\bar \chi}  {\mathbf H}_{t,T}(\Theta_t),
	    \end{align}
	    with a probability that exceeds $1/4$
	    
	    \subsection*{Appendix D: Proof of Theorem \ref{thm:perf}}
	    
	Combining \eqref{eq:spectralbound} and the definition of $\rho_v$, we find that the following inequality with probability at least $1/4$ holds. 
	        \begin{align}\label{eq:somebound5}
	        \rho_v(\mathbf H_{t,T}(\Phi_t))   \Sp &\leq  \Sp \dfrac {4  \bar \chi}{1-\epsilon} \Sp \Sp \mathrm{Tr}(\mP_{t,T}(\Theta_t)) \\
	        & =\dfrac {4 \bar \chi}{1-\epsilon} \Sp \rho_v(\mathbf H_{t,T}(\Theta_t)) .
	        \end{align}
	        The proof of \eqref{eq:comega_bound_33} is similar. For the entropy estimation measure,  combining \eqref{eq:spectralbound} and definition of $\rho_e$, we find that the following inequality with probability at least $1/4$ holds. 
	        \begin{align*}
	        \rho_e(\mathbf H_{t,T}(\Phi_t)) &=  \log(\det (\Omg_{t,T}(\Phi_t)^{-1})) \\
	        & \leq  \log\left (\det\left (\dfrac{4\bar \chi}{1-\epsilon} \Omg_{t,T}(\Theta_t) ^{-1}\right ) \right ) \\
	        &= n\log \left ( \dfrac{4\bar \chi}{1-\epsilon} \right ) +\rho_e(\mathbf H_{t,T}(\Theta_t)).
	        \end{align*}



		
	\bibliographystyle{IEEEtran}
	
	\bibliography{mybib} 
	
\end{document}